\documentclass[twoside,11pt]{article}

\usepackage{amsmath}
\usepackage{mathrsfs}
\usepackage{algorithm}
\usepackage{algpseudocodex}
\algdef{SE}{Begin}{End}{\textbf{begin}}{\textbf{end}}
\usepackage[utf8]{inputenc}
\usepackage{graphicx}
\usepackage{comment}

\usepackage{jmlr2e}

\DeclareMathOperator*{\argmin}{argmin}
\DeclareMathOperator*{\argmax}{argmax}
\DeclareMathOperator*{\var}{var}

\usepackage{lastpage}

\firstpageno{1}

\begin{document}

\title{Optimal Initialization of Batch Bayesian Optimization}

\author{\name Jiuge Ren  \email jren@mail.yu.edu \\
       \addr Department of Computer Science and Engineering\\
       Yeshiva University\\
       New York, NY 10016, USA
       \AND
       \name David Sweet \email david.sweet@yu.edu \\
       \addr Department of Computer Science and Engineering\\
       Yeshiva University\\
       New York, NY 10016, USA}

\editor{}

\maketitle

\begin{abstract}

    Field experiments and computer simulations are effective but time-consuming methods of measuring the quality of engineered systems at different settings. To reduce the total time required, experimenters may employ Bayesian optimization, which is parsimonious with measurements, and take measurements of multiple settings simultaneously, in a \textit{batch}. In practice, experimenters use very few batches, thus, it is imperative that each batch be as informative as possible. Typically, the initial batch in a Batch Bayesian Optimization (BBO) is constructed from a quasi-random sample of settings values. We propose a batch-design acquisition function, Minimal Terminal Variance (MTV), that designs a batch by optimization rather than random sampling. MTV adapts a design criterion function from Design of Experiments, called I-Optimality, which minimizes the variance of the \textit{post-evaluation} estimates of quality, integrated over the entire space of settings. MTV weights the integral by the probability that a setting is optimal, making it able to design not only an initial batch but all subsequent batches, as well. Applicability to both initialization and subsequent batches is novel among acquisition functions. Numerical experiments on test functions and simulators show that MTV compares favorably to other BBO methods.
\end{abstract}


\section{Introduction}

Batch Bayesian optimization (BBO) designs a short sequence of experiments, each consisting of a set of parallel measurements (a \textit{batch}). Measurements may be in the form of (i) field experiments or (ii) computer simulations, in subject areas such as recommender systems \citep{fb}, materials science \citep{ms,ms2}, A/B testing of applications at technology companies \citep{ab,e4e}, or automotive or aerospace design \citep{as}. The aim of the experiments is to find a good configuration of the system under study, answering questions such as: Which version, A or B, of our software produces more user engagement \citep{ab}? Which "recipe" will produce a metallic glass \citep{ms2}?

Each batch may take a long time to complete. One to two weeks is typical for online A/B testing \citep{ab,ab2}. One day might be appropriate for a materials science experiment \citep{ms2}. Simulations may take a day \citep{as} to weeks \citep{ms} to complete. As such, experimental BBO often consists of few batches, each with few to many measurements. In the literature we find examples of 2-5 batches (2 batches in \citeauthor{fb}, \citeyear{fb}; 3 in \citeauthor{ms2}, \citeyear{ms2} and \citeauthor{fb3}, \citeyear{fb3}; and 5 in \citeauthor{ms}, \citeyear{ms}) containing 3-50 measurements (3 measurements in \citeauthor{fb}, \citeyear{fb}; 20 in \citeauthor{fb3}, \citeyear{fb3}; 25 in \citeauthor{abm}, \citeyear{abm}; 31 in \citeauthor{fb}, \citeyear{fb}; and 50 in \citeauthor{abm}, \citeyear{abm}). In Bayesian optimization parlance, we call the limit on the number of rounds the \textit{budget}, and limit on the number of measurements/round the \textit{capacity}. We consider the scenarios just discussed to have a low budget and small-to-moderate capacity.

We claim that the initial batch in a BBO sequence is important yet under-studied. It is important because the initial batch contains a significant portion of the total number measurements taken across all batches, due to the low budget. Also, since each subsequent, improvement-batch design will depend on the initialization measurements, better initialization will indirectly improve those designs. Research \citep{rec,fb,q,bts} typically focuses on how to design the improvement batches, not on how to initialize. We emphasize that initialization is acquisition. The first batch is as expensive as any other and, thus, is worthy of greater attention in both research and in practice.

The initial batch is distinct from improvement batches in that it must be designed \textit{de novo}---with no prior measurements. Lacking measurements to inform the design, the initial batch is typically chosen randomly or quasi-randomly \citep{rec,scbo,fb,fb3}, ex., from a Sobol' sequence \citep{exp}. For implementation examples, see \texttt{gp\_minimize} in scikit-optimize \citep{skopt}, Meta's Ax \citep{ax}, or dragonfly \citep{dragonfly-code, dragonfly-paper}.

We propose a batch design method, which we call Minimal Terminal Variance (MTV), that generates an initial batch by optimizing an acquisition function. Further, we use the same acquisition function to design all batches in a BBO sequence. MTV is competitive with other BBO methods and is the only method that (i) initializes BBO via optimization rather than random sampling, and (ii) may be effectively applied to both initialization and improvement batches.

\subsection{Background}

We model the system under study---whether simulated or physical---as a black-box function, $f(x)$, that represents the quality of the system when run at a parameter setting, $x$. An experiment measures $y = f(x) +  \varepsilon$, the quality corrupted by some noise. Our goal is to find the optimal parameter setting

\[
x_* = \argmax_{x \in \mathcal{X}} f(x).
\]

The main challenge of experimental optimization \citep{nist} is that experiments are \textit{expensive} (ex., in dollars, time, or risk). To minimize the expense, we optimize a model of the system under study, rather than the system itself. The model, called a \textit{surrogate}, estimates $f(x)$. Querying it serves as a cheap proxy for running an experiment. The optimal $x$ of a surrogate is called an \textit{arm}, and it becomes a candidate for experimental evaluation. In the optimal design of experiments (DoE) setting, the surrogate is a polynomial in $x$. In Bayesian optimization it is typically a Gaussian process (GP).

A GP associates a multivariate normal distribution, $\hat{y}(x_i) \sim \mathcal{N}(\mu(x_i | \mathcal{D}), \Sigma(x_i | \mathcal{D}))$, to any set of parameters, $x_i$, conditional on the measurements collected so far, $\mathcal{D} = \{x_m, y_m\}$ \citep{e4e}, where

\begin{equation}
\begin{aligned}
\mu(x_i) &= K(x_m,x_i)^T (K(x_m,x_m) + \sigma^2I)^{-1} y_m  \\
\Sigma(x_i) &= K(x_i, x_i) - K(x_m, x_i)^T (K(x_m,x_m)+ \sigma^2I)^{-1} K(x_m,x_i).
\end{aligned}
\label{eq:GP}
\end{equation}
Specifying the form of the covariance matrices $K(\cdot,\cdot)$ is part of the modeling process. The measurement noise level is $\sigma^2 = \var \varepsilon$. We'll denote the model, Equation~\eqref{eq:GP}, by $\mathcal{GP}(x)$ or $\mathcal{GP}(x|\mathcal{D})$ to make explicit that it is conditioned on the collected data.

Note the following properties of these equations:
\begin{description}
    \item[Property 1] $\mathcal{GP}(x)$ jointly models the expectation, $f(x)$, the measurement (aleatoric) uncertainty, $\sigma^2$, and the model (epistemic) uncertainty, $K(\cdot,\cdot)$.
    \item[Property 2] A single draw from the joint distribution, Equations~\eqref{eq:GP}, would deliver values, $y_i$, lying on a smooth surface over $x$, where the smoothness is determined by $K(\cdot,\cdot)$. It can be useful to imagine a joint draw as a two part process: Randomly draw a single, smooth function that is consistent with the GP, then sample value $y_i$ from that function at locations $x_i$.
    \item[Property 3] The covariance depends only on the parameters, $x_m$, where measurements have been taken and not on the values measured, $y_m$.
\end{description}
We'll refer to these properties in later discussion.

To propose arms, optimal DoE and Bayesian optimization methods may optimize some function of $\mu(x_i)$ and $\Sigma(x_i)$, that, generally speaking, is designed to reduce the total number of measurements needed to find $x_*$. In Bayesian optimization such functions are called \textit{acquisition functions}.

\subsection{Related Work}

Our work studies batching methods with a focus on the initialization batch. There are a few common approaches to initialization and several approaches to batching in the literature. We discuss them here.

\textbf{Initialization} Initial measurements are generally taken at random parameter values, according to a space-filling design, or by Latin hypercube design \citep{hb,dace,fl}, as seen both in the literature \citep{ego,rec,scbo,fb3} and in popular software packages \citep{ax,skopt}. These approaches to initialization are inherited from the design and analysis of computer experiments \citep{dace}. They are motivated by the rationale that since GP surrogates are flexible, in the absence of pre-existing measurements one has no knowledge of the form the GP will take \citep{hb}, therefore it is best to take measurements at several points that are spread out in the parameter space. Different notions of "spread out" produce different initial designs \citep{dace}. As an alternative to randomization, we produce the initial batch by optimizing an acquisition function, MTV.

Common approaches to selecting multiple arms for a batch employ random sampling, greedy optimization (one arm at a time), and joint optimization (all arms simultaneously). We describe a few examples of methods here and study them in Section \ref{sec:numerical-experiments}. For each method, one must first run an initialization batch and use the resulting measurements, $\mathcal{D}$, to construct a surrogate, $\mathcal{GP}(x|\mathcal{D})$. 

\begin{description}
\item[Thompson Sampling (TS)] Denote a parameter by $x$ and the probability that $x$ maximizes $f(x)$ as $p_*(x)$. Sample an arm $x_a \sim p_*(x)$ by computing $x_{ts} = \argmax_{x_s} \hat{y}(\{x_s\})$, where the points $x_s$ are taken from a uniform distribution (or from a Sobol' sequence), and $\hat{y}(\{x_s\})$ is a joint draw at all $x_s$ from the GP. By Property 2, a TS approximates a maximization of a single function drawn from $\mathcal{GP}(x|\mathcal{D})$. Create a batch of size $B$ by repeatedly sampling ($B$ times) in this way.

\item[Determinantal Point Process Thompson Sampling (DPP-TS)] Since the samples in a TS batch are drawn independently, they may cluster together, which is inefficient for exploring the parameter space. DPP-TS reduces clustering by drawing a $B$-element sample of arms $x_a \sim D(\{x_a\})\Pi_a p_*(x_a)$, where $D(\{x_a\})$ is a repulsion term that is larger when the batch $\{x_a\}$ is more spread out \citep{dpp}.

\item[GIBBON] GIBBON, among other generalizations \citep{gibbon}, extends max-value entropy search \citep{mves} from sequential to batch optimization. The sequential acquisition function is the mutual information, $MI(x_a, y_*)$, between a proposed arm and the maximum measured objective value, $y_*$. To create a batch, GIBBON first finds a single arm, $x_1$, by maximizing $MI(x_1, y_*)$, then finds each additional arm greedily by maximizing $MI(\{x_1 \dots x_k\}, y_*)$ with all arms but $x_k$ held fixed.

\item[q-Simple Regret] qSR jointly optimizes a batch of $q=B$ arms, by maximizing the acquisition function $\int_\mathcal{Y} dy_q  \max y_q$, where the space $\mathcal{Y}$ is the $q$-dimensional space of all possible joint measurements of the vector $y_q$. The $\max$ (in the integral) is taken over the $q$ dimensions. The integral is approximated with Monte Carlo sampling and the reparameterization trick \citep{q}. Replacing $y_q$ with either the upper confidence bound \citep{ucb} or expected improvement \citep{ego} acquisition function constructs the batch method q-Upper Confidence Bound or q-Expected Improvement, respectively \citep{q}.
\end{description}

Note that all of these batch methods are applied to improvement batches only and require the prior, independent construction of an initialization batch, ex., arms chosen from a Sobol' sequence. By contrast, MTV does not require prior initialization.

\section{Method: Minimal Terminal Variance (MTV)}

\newcommand{\dd}{\mathop{}\!{d}}

MTV designs a batch of arms by minimizing the terminal (i.e., post-measurement) prediction variance, averaged over the parameter space, $\mathscr{X}$, and weighted by the probability that $x \in \mathscr{X}$ is the maximizer:
\begin{equation}
\label{eq:int}
MTV(x_a) = \int_\mathscr{X} \dd{x} \ p_*(x) \sigma^2(x|x_a)
\end{equation}
where $\sigma^2(x|x_a)$ is the variance estimate given by the surrogate, $\mathcal{GP}(x|x_a)$, conditioned on measurement of a batch of arms, $x_a$. If we were to omit the weighting, $p_*(x)$, the integral would be called I-Optimality \citep{iopt} in the DoE literature. With a non-constant weight, $p(x)$, the integral is called Elastic I-Optimality by \citet{eiopt}. We refer to the specific case where the weight is $p(x) = p_*(x)$ as MTV.

The batch design is given by the joint optimization of all $B$ arms, $x_a$,
$$x_a = \argmin_{x_a^\prime} MTV(x_a^\prime).$$
The same design procedure is used for every round---whether for initialization or improvement.

To perform a practical calculation, we replace the integral with the approximation
$$MTV(x_a) \approx \sum_{i} \sigma^2(x_i|x_a)$$
where the weighting is effected by sampling the evaluation points $x_i \sim p_*(x)$ and the overall constant is ignored as it does not affect the result of the optimization (i.e., the $\argmin$).

\subsection{Sampling from \texorpdfstring{$p_*(x)$}{p-star}}

To generate samples $x_i \sim p_*(x)$, we adapt the Metropolis algorithm \citep[Chapter~1]{mcmc} to create a problem-specific MCMC sampler. In general, to sample via Metropolis from a distribution, $p(x)$, we choose a random starting value for $x$, then (i) propose a perturbation, $x^\prime = x + \sim\mathcal{N}(0, \varepsilon^2)$, (ii) calculate the \textit{acceptance probability}, $\alpha = \min(1, p(x^\prime) / p(x))$, then (iii) with probability $\alpha$, replace $x$ by $x^\prime$. Then we record $x$. To generate another sample from $p(x)$, we repeat the steps above starting from the recorded $x$ value.

Since the problems we solve are bounded in $x$ (generally, $x \in [0,1]^d$), the perturbation might not lie within the bounds. To correct for this, we replace the simple perturbation with a hit-and-run perturbation \citep[Chapter~6]{mcmc}, in which we (i) choose a random direction, $\delta$, (ii) find the distances, $\lambda_+$ and $ \lambda_-$, from $x$ to the boundary along $\delta$ and $-\delta$, then (iii) set $x^\prime = x + \eta \delta$ with $\eta \sim \mathcal{TN}(0, \varepsilon^2; -\lambda_-, \lambda_+)$, where $\mathcal{TN}$ is a 1D normal distribution truncated to $-\lambda_-, \lambda_+$. 

To sample from $p_*(x)$, we need to evaluate $\alpha$, which requires us to estimate $p_*(x^\prime)$ and $p_*(x)$. To do this, we could take $n$ joint (Property 2) samples, $[y(x), y(x^\prime)] \sim \mathcal{GP}([x,x^\prime])$ then estimate $p_*(x) = \text{count}(y(x) > y(x^\prime)) /n$, and similarly for $p_*(x^\prime)$. In practice, however, we set $n=1$ then take $\alpha = 1$ if $\mu(x^\prime) > \mu(x)$ or $\alpha=0$ otherwise. Importantly, since the time to sample from $\mathcal{GP}(x)$ scales with $n^3$ \citep{tut}, using $n=1$ uses considerably less computation time (per update of the MCMC sampler) than would the construction of a more precise estimate of $\alpha$ (i.e., with $n$ large).

A sequence of $x$ values sampled via MCMC is autocorrelated \citep{mcmcintro} and, thus, the entire sequence of samples, viewed collectively, might deviate from a (desired) set of independent samples from the target distribution, $p(x)$. Also, it takes several iterations before the samples accurately reflect the target distribution (the burn-in period, \citeauthor{mcmcintro}, \citeyear{mcmcintro}). In light of these facts, we iterate several ($M$) times and only record the final $x$ value. To obtain $N_x$ samples, we run $N_x$ samplers in parallel and record the $N_x$ final $x$ values.

As the number of measurements available to the surrogate increases, $p_*(x)$ will become increasingly concentrated around $x_* = \argmax_x \mu(x)$. Thus, for efficiency, we initialize the sampler at $x_*$.

Finally, since the width of $p_*(x)$ varies from round to round, there is no obvious single choice for the value of $\varepsilon$, so we adapt $\varepsilon$. If too few (resp., many) of the $N_x$ values (of $x$) get updated in an iteration of the sampler, we shrink (resp., grow) $\varepsilon$. The $p_*(x)$ sampling algorithm is summarized in Algorithm~\ref{alg:mcmc}.

Note that when we prepare the initialization batch of a Bayesian optimization---when no measurements are yet available---we assert (as our prior) that $p_*(x)$ is uniform over $\mathcal{X}$. In that case, in the interest of execution speed, our code draws $x_i$ from a Sobol' sequence, although Algorithm~\ref{alg:mcmc} would produce an equally-valid result.

\begin{algorithm}
\caption{P-Star Sampler}
\label{alg:mcmc}
\renewcommand{\thealgorithm}{}
\floatname{algorithm}{}
\begin{algorithmic}[1]
    \State $x_* = \argmax_{x^\prime} \mu(x^\prime)$ \Comment{$\mu(x^\prime)$ is mean of $\mathcal{GP}(x^\prime)$}
    \State $x_i = x_*$, for $i \in 1...N_x$ \Comment{Set all $x_i$ to same value}
    \ForAll{$m \in 1, \dots, M$}
    \State $x^\prime_i = \texttt{HNR}(x_i)$ \Comment{Perturb $x_i$ via Hit-and-Run}
    \State $[ y_i, y^\prime_i ] \sim \mathcal{GP}([x_i, x^\prime_i])$ \Comment{Joint sample}
    \State $x_i \leftarrow x^\prime_i$ \textbf{if} $y^\prime_i > y_i$ \Comment{Update selected samples}

    \EndFor
    \State \Return $x_i$ \Comment{$N_x$ samples from $p_*(x)$}

\end{algorithmic}\end{algorithm}

\subsection{Minimizing the integral approximation}

Once the $x_i$ are sampled, they are held fixed while $MTV(x_a) \approx \sum_{i} \sigma^2(x_i|x_a)$ is optimized (minimized). We find that if we begin the optimization at candidate arms chosen from $x_i$, the optimizer finds lower (more optimal) values of $MTV(x_a)$ than if we begin at values of $x$ randomly sampled from $\mathcal{X}$ (which is the default behavior of the acquisition function optimizer in, ex., \texttt{BoTorch}, \citeauthor{botorch_code}, \citeyear{botorch_code}). The importance of careful initialization of the acquisition function optimizer was studied in \citet{afic}. We study the impact of this decision on MTV in Section \ref{sec:ablations}.

It is (fortunately) possible to efficiently estimate $\sigma^2(x_i|x_a)$ in the inner loop of the acquisition function optimization. First, note that the covariance of $\mathcal{GP}(x|x_a)$ depends only on the parameter values, $x$ and $x_a$ (Property 3); specifically, it does not depend on the measurement outcomes, $y(x_a)$. Therefore, we can calculate $\sigma^2(x_i|x_a)$ without running an experiment.

Second, one can condition a GP on $x_a$ quickly, without refitting \citep{es}. This is sometimes called "fantasizing". In each iteration of the acquisition function optimization we fantasize a measurement at $x_a$ and the $\mathcal{GP}(x|x_a)$ that would result from it. From the fantasized $\mathcal{GP}(x|x_a)$ we calculate $\sigma^2(x_i|x_a)$. Fantasized GPs are used in entropy search \citep{es} and in non-myopic \citep{nonmyopic} acquisition methods. Unlike MTV, those methods also consider possible future outcomes of $y(x_a)$ as draws from the conditioned GP. We rely on the support for conditioned/fantasized models in the \texttt{BoTorch} \citep{botorch_paper} and \texttt{GPyTorch} \citep{gpytorch} libraries.

\begin{algorithm}
\caption{Minimal Terminal Variance (MTV) Batch Design}
\label{alg:MTV}
\renewcommand{\thealgorithm}{}
\floatname{algorithm}{}
\begin{algorithmic}[1]
     \State $x_i$ = \Call{P-Star Sampler}{ } \Comment{$x_i \sim p_*(x)$}
     \State Choose $x_a \subseteq x_i$ \Comment{$B$ arms, $x_a$}
    \Repeat
     \State Propose $x_a^\prime \in X$ \Comment{Ex., by L-BFGS-B}
     \State Condition $\mathcal{GP}(x|x_a^\prime)$ \Comment{\citet{es}}
     \State Calculate $MTV(x_a^\prime) \approx \sum_{i} \sigma^2(x_i|x_a^\prime)$ \Comment{Approximate Equation \eqref{eq:int}}
     \State $x_a \leftarrow x_a^\prime$ \textbf{if} $MTV(x_a^\prime) < MTV(x_a)$ \Comment{Track minimum}
    \Until{Converged}
    \State \Return $x_a$  \Comment{$x_a = \argmin_{x_a^\prime} MTV(x_a^\prime)$}

\end{algorithmic}\end{algorithm}

The batch design procedure is presented in Algorithm~\ref{alg:MTV}. A Python implementation is available as \texttt{AcqMTV}
in our Github repository, \url{https://github.com/dsweet99/bbo}. Note that to optimize MTV we use L-BFGS-B, via \texttt{BoTorch}'s function \texttt{optimize\_acqf}, which, in turn, calls \texttt{scipy.optimize.minimize} \citep{scipy}.

\begin{figure}
   \centering
   \includegraphics[width=0.5\linewidth]{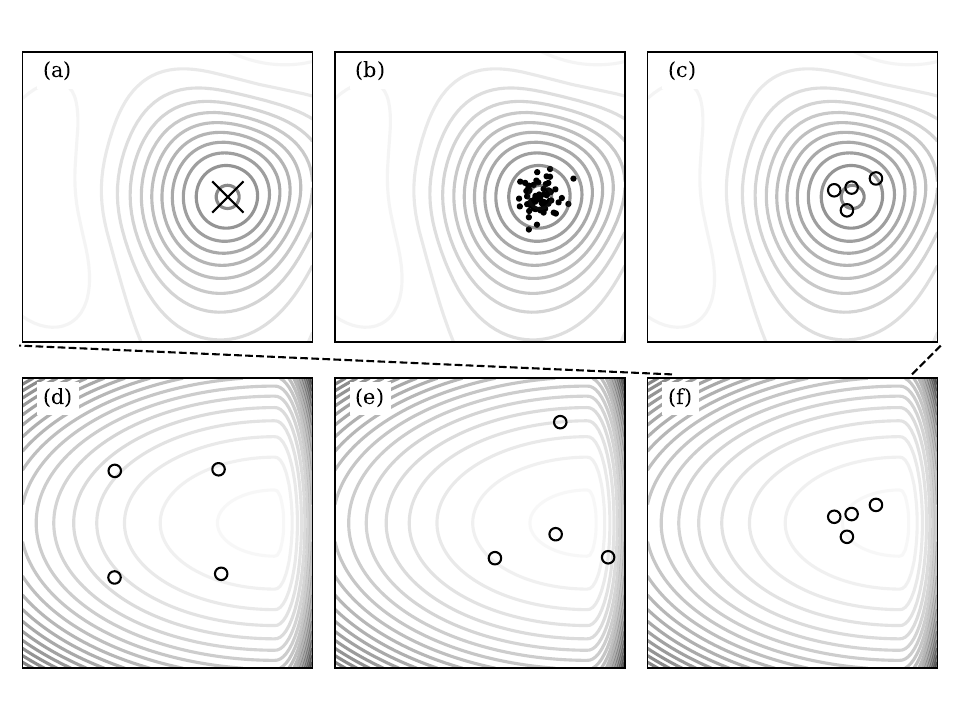}\hfill
    \caption{
    The top row shows the construction of a single MTV batch: (a) Find the maximizer (marked X) of $\mu(x)$, the GP mean. Contour lines show $\mu(x)$. (b) Draw $10B=40$ points, $x_i \sim p_*(x)$, via Algorithm~\ref{alg:mcmc}. (c) Jointly determine the $B=4$ arms, $x_a$, that minimize (over $x_a$) the mean (over $x_i$) terminal GP variance, $\sigma^2(x|x_a)$, via Algorithm~\ref{alg:MTV}. \\
    The bottom row shows the batches of arms chosen at: (d) round zero, initialization, and improvement rounds (e) one and (f) two (the round in the top row). Contour lines show values of the true function, $f(x)$.
    }

    \label{fig:explain}
\end{figure}

The top row of Figure~\ref{fig:explain} depicts the design of a single batch. We first follow Algorithm~\ref{alg:mcmc} and (a) find the maximizer of the surrogate mean, $x_* = \argmax_x \mu(x)$, then, (b) draw many samples (eg., $N_x = 10B$), $x_i \sim p_*(x)$. Finally we (c) use Algorithm~\ref{alg:MTV} to minimize $MTV(x_a)$, finding $B=4$ arms. The bottom row shows the arm locations at rounds 0, 1, and 2. Notice that the arms collect around the maximum of the function, near the samples $x_i \sim p_*(x)$, but are not, themselves, samples from $p_*(x)$.

\section{Numerical experiments}
\label{sec:numerical-experiments}
To evaluate MTV we run batch Bayesian optimizations taking, as the metric of quality of an optimization, the maximum measured value, $y = f(x) + \varepsilon$, over all rounds. As stand-ins for the system/simulation, we use nine common optimizer test functions and two reinforcement learning tasks, MountainCarContinuous-v0, and Hopper-v4.

We match the strict requirements of field-experimental optimization or the optimization of long-running simulations in that we (i) keep a low budget, i.e. we run only three rounds, and (ii) we assume moderate capacity, ie. each round is a batch of moderate size -- 4-30 arms.

In addition to comparison of performance with other BBO methods (Sections~\ref{sec.performance-comparison} and ~\ref{sec:rlsim}), we also look more closely at the initialization batch (Section~\ref{sec:initialization}) and show that MTV finds better parameter settings than standard methods even when no prior measurements are available. To test the construction of the MTV method we run ablations, removing each of the three crucial components--Algorithm~\ref{alg:mcmc}, Algorithm~\ref{alg:MTV}, and non-random initialization of the acquisition function optimizer--and show that each is valuable to the method. Finally, in Section~\ref{sec:ensembling}, we explore an approach that is popular with practitioners of combining BO methods to solve specific problems and find that MTV can combine well with existing methods.

\subsection{Performance comparison}
\label{sec.performance-comparison}
We first compare MTV to other BBO methods by optimizing test functions (see \citeauthor{optfun}, \citeyear{optfun}). Some functions (ex., Ackley) have their optima at the center-point parameter, $x=0$, meaning that any method that initially guesses the center point will have an unfair advantage in the comparison \citep{centerbias}. To mitigate this center bias, we move the center point to a uniformly-randomly chosen value, $x_0$, and distort the space so that the boundaries at $0$ and $1$ remain fixed. Along each dimension, $d$, we distort like this
$$x^\prime =
\begin{cases}
\frac{x - x_0}{1 + x_0} & x < x_0\\
\frac{x - x_0}{1 - x_0} & x > x_0
\end{cases}
$$
The value $x_0$ is set at the beginning of an optimization run. We apply this transformation to all functions, irrespective of where maximum lies.

We define a \textit{problem} as the pair (\textit{function name}, $x_0$), where \textit{function name} is one of nine test functions when the parameter space dimension is $d > 1$ and one of seven functions when $d=1$. (We use only seven functions when $d=1$ because two functions, Michalewicz and
Rosenbrock, are not defined when $d=1$.)

\begin{figure}[bt!]
   \centering
   \includegraphics[width=0.75\linewidth]{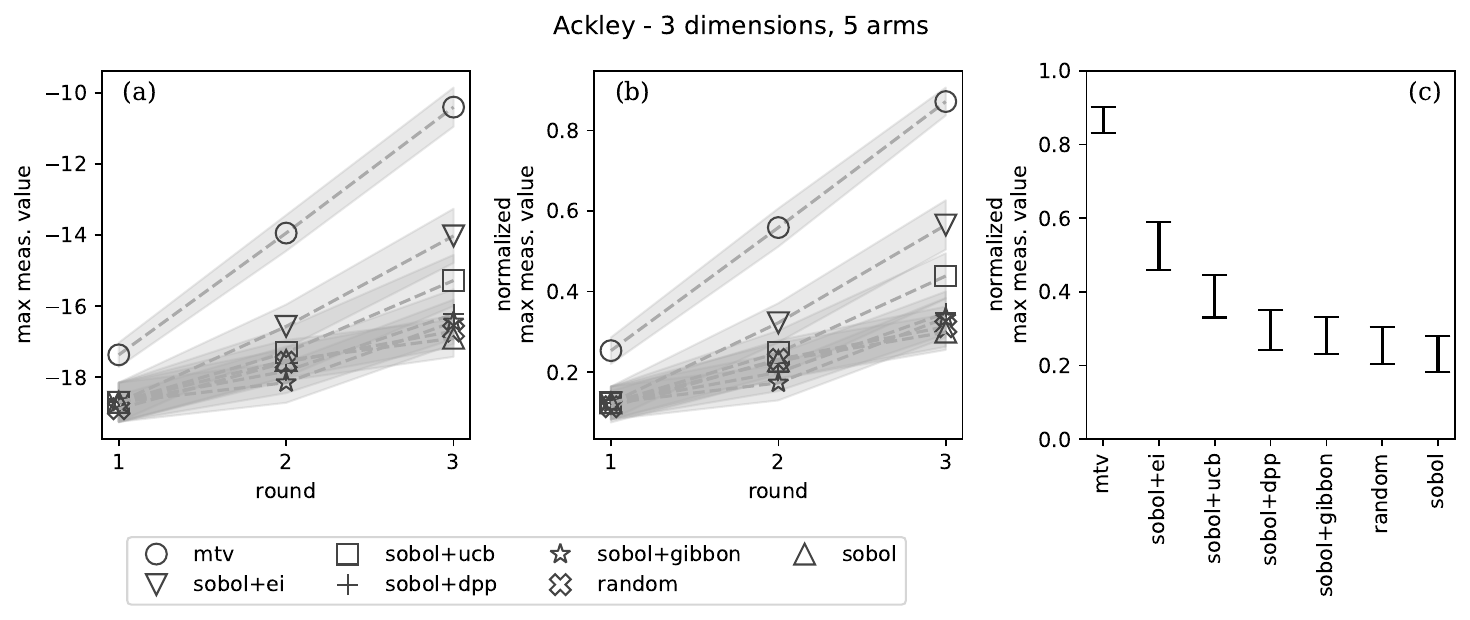}\hfill
    \caption{Comparison of MTV to other batch Bayesian optimization methods on the 3D Ackley function. (a) Results are averaged over 30 runs.  Each run randomly distorts the Ackley function to mitigate center bias (see Section~\ref{sec.performance-comparison}). (b) Results are range-normalized across optimization methods before averaging. (c) The final-round results are summarized.
    }
    \label{fig:ackley}
\end{figure}

We optimize a set of multiple problems with a common \textit{function name} using each of several methods, and average the results across problems. For example, Figure~\ref{fig:ackley} shows 30 problems generated with the 3D Ackley function and optimized by several BBO methods:
\begin{itemize}
\item \texttt{sobol+sr}, q-Simple Regret \citep{reparam},
\item \texttt{sobol+ucb}, q-Upper Confidence Bound \citep{reparam},
\item \texttt{sobol+ei}, q-Noisy Expected Improvement \citep{qnei},
\item \texttt{sobol+gibbon}, GIBBON with greedy batches \citep{gibbon}, and
\item \texttt{sobol+dpp}, DPP-TS \citep{dpp}.
\end{itemize}
In each method, the first round's arms are generated via a Sobol' (quasi-random, space-filling) sequence \citep{hb,dace}. As baselines for comparison, we also include methods \texttt{random}, which samples uniformly in $\mathcal{X}$, and \texttt{sobol}, which also samples uniformly but with low discrepancy, using a Sobol' sequence \citep{hb}.

In Figure~\ref{fig:ackley}(b), before averaging we normalize the results to the range of $y$-axis values produced for each problem over all methods being compared. For each run, the $y$-axis value is the maximum of the measured test function values so far in the run. Figure~\ref{fig:ackley}(c) summarizes the normalized values from the final round, ordering them from highest to lowest.

For all methods, except DPP-TS, we use implementations of acquisition functions from the \texttt{BoTorch} library \citep{botorch_paper, botorch_code}. A DPP-TS implementation was provided to us by the authors of \citet{dpp} and also was published with \citet{sober}. Code to reproduce these experiments is available in \url{https://github.com/dsweet99/bbo}.

\begin{figure}[bt!]
   \centering
   \includegraphics[width=1.0\linewidth]{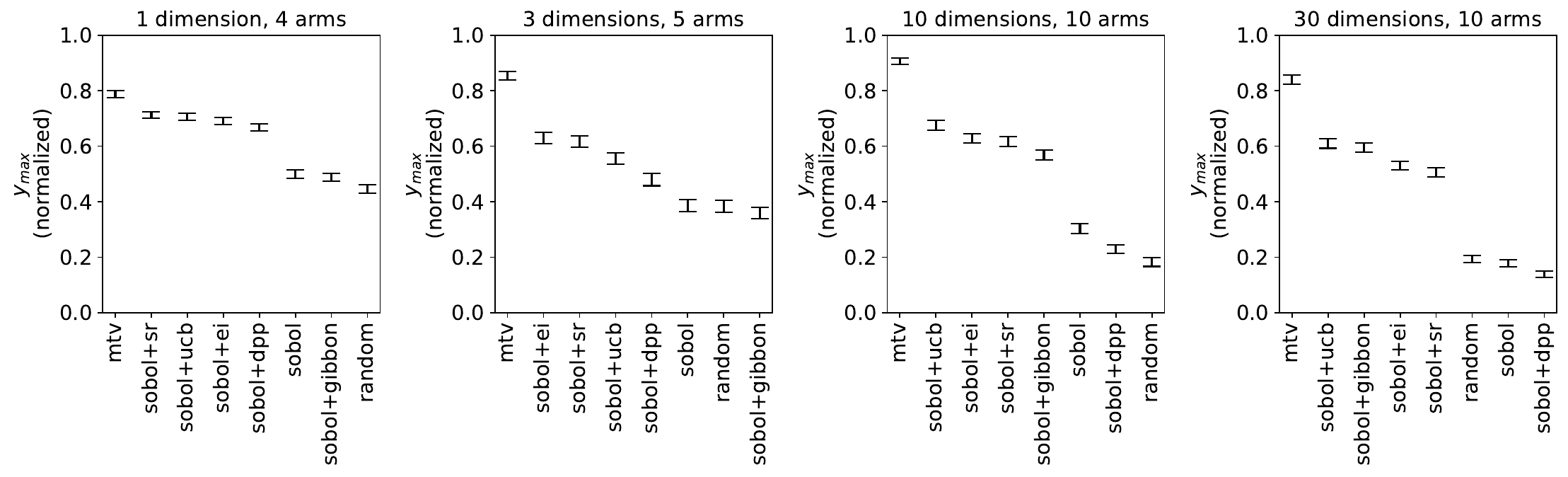}\hfill
    \caption{
    Comparison of MTV to several other acquisition functions and two baselines (\texttt{sobol} and \texttt{random}).
    }
    \label{fig:compare}
\end{figure}

Figure~\ref{fig:compare} aggregates results over all applicable test functions in dimensions 1, 3, 10, and 30, in the style of Figure~\ref{fig:ackley}(c). Each point is an average over all problems, over 30 runs of each problem, after range normalization.

MTV outperforms other batch methods across a large range (1D to 30D) of problem dimension. We note that while \texttt{mtv} is always in the top position, the ranking, otherwise, is not stable with dimension. For example, there is no consistent "second place" or "worst-ranked" method.

\subsection{Reinforcement Learning Simulators}
\label{sec:rlsim}

We continue the comparisons with other BBO methods, now using reinforcement learning (RL) simulators from Gymnasium \citep{gynasium}, formerly OpenAI gym. The RL simulators (i) provide functional forms for $f(x)$ arising from simulated physics and (ii) add uncertainty to the measurements. These aspects may make the RL problems more like real-world simulations or field experiments than the test problems of section \ref{sec.performance-comparison}.

\subsubsection{MountainCarContinuous-v0}
\begin{figure}[bt!]
   \centering
   \includegraphics[width=0.75\linewidth]{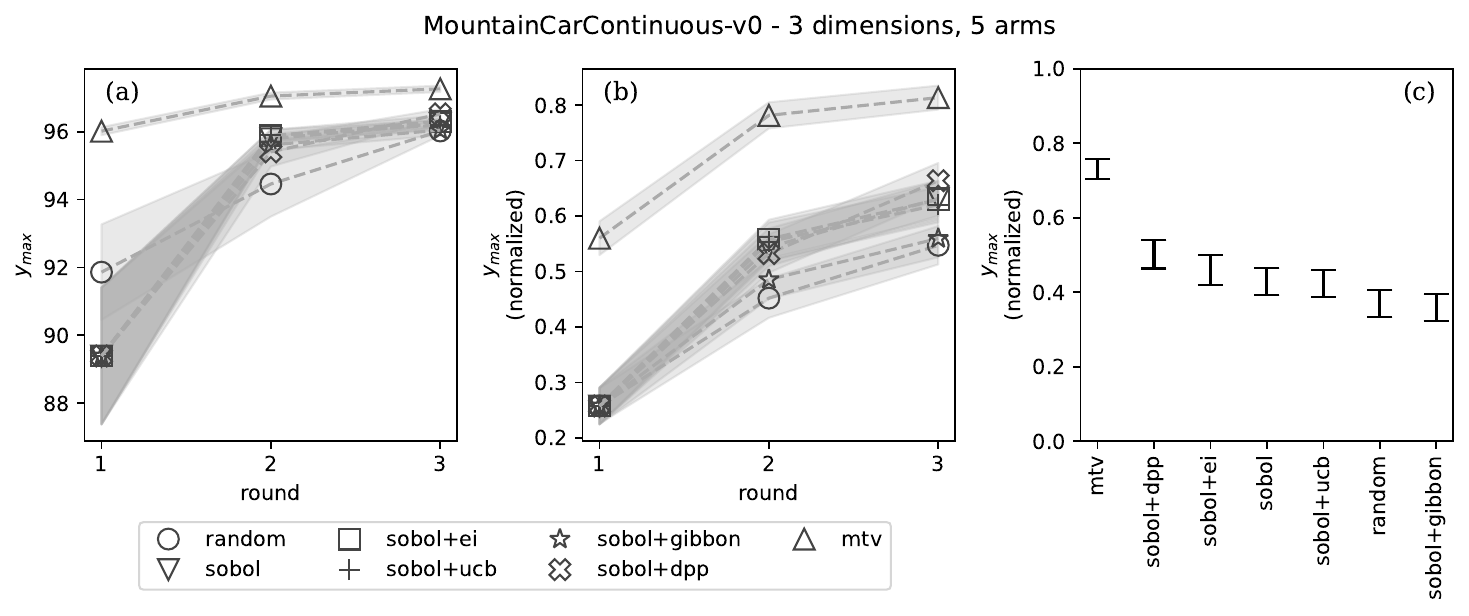}\hfill
    \caption{
    Optimizing a 3-dimensional controller for the mountain car RL simulator with 5 arms/batch. We run 100 replications to calculate error bars. (a) The y axis shows the return (total episode reward). (b) The y axis is range-normalized across optimization methods. (c) The range-normalized value of the final round.
    }

    \label{fig:mcc}
\end{figure}

The mountain car simulator \citep{mcc_orig} presents the task of optimizing a controller to move an underpowered car out of a valley to the top of a mountain, which we call the goal \citep{mcc_orig}. Accelerating the car yields a negative reward, and the only positive reward is received when the car reaches the goal. Such a problem is a challenge \citep{mcc_dqn} for an optimizer that uses only local (e.g., gradient) information, as the reward gradient always points toward controllers that accelerate less (to reduce the negative reward of using fuel), but without acceleration the car cannot reach the goal. Bayesian optimizers use global information and, so, are able to solve this problem. Figure~\ref{fig:mcc} shows that MTV finds a goal-reaching controller more often than the other BBO methods.

\texttt{MountainCarContinuous-v0} has a two-dimensional state space and a one-dimensional action space. We implemented a controller based on the linear controllers in \citet{rl_linear}. Our controller maps a state vector to an action vector with
$$a = k B s$$
where $B$ is a $\dim(s) \times \dim(a)$ matrix of parameters and $k$ is an overall scale parameter. The state is normalized (elementwise) dynamically with its running average, $\mu$, and running standard deviation, $\sigma$,
$$ s = (s - \mu) / \sigma$$
The controller has $1 + \dim(s) \times \dim(a) = 1 + 2 \times 1 = 3$
 parameters.

Our tests run three rounds of simulations with 5 arms/batch. The measurement, $y(x)$, is the sum of all of the rewards---one per time step---of the simulation, also called the \textit{return}. Each time the simulator starts, the mountain car is given a randomly-chosen position and velocity, leading to noise in the measurement: $y(x)=f(x) + \text{noise}$. To reduce this noise, we run the simulator 30 times and take the average return as the measurement, $y$

\subsubsection{Hopper-v4}
\begin{figure}[bt!]
   \centering
   \includegraphics[width=0.75\linewidth]{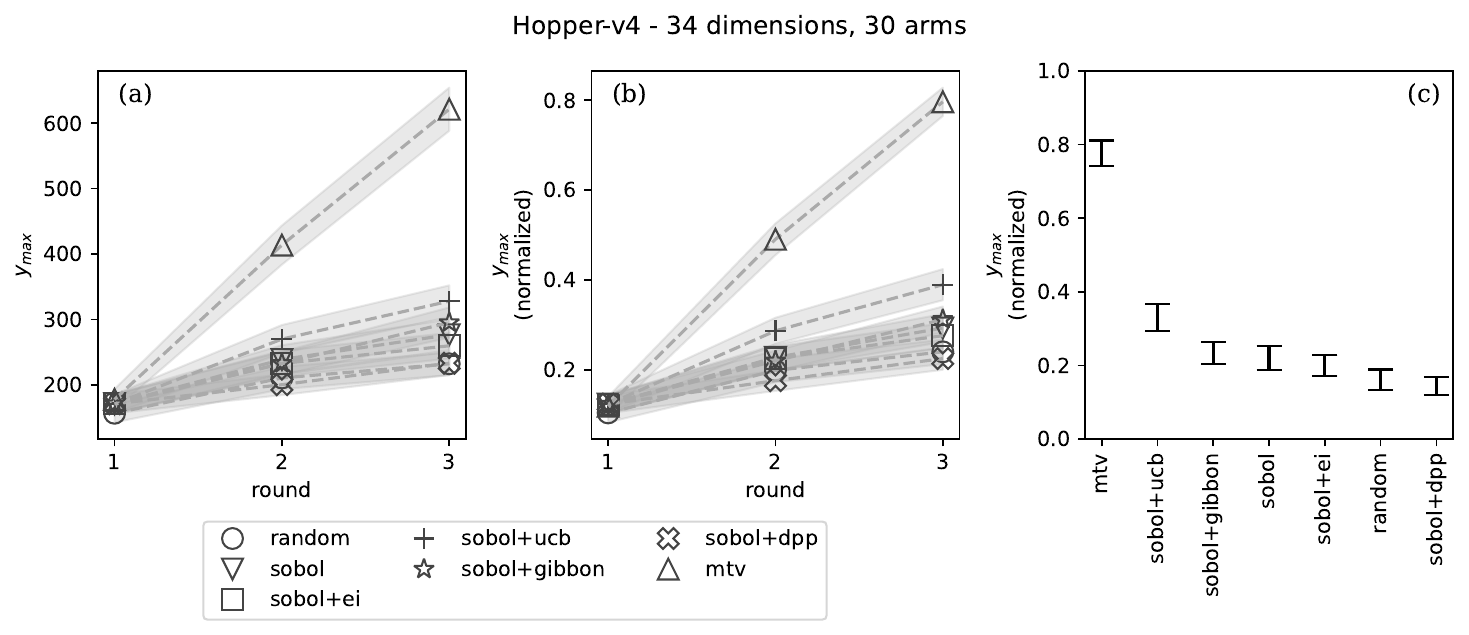}\hfill
    \caption{
   Optimizing a 34-dimensional controller for the a one-legged hopping robot RL simulator with 30 arms / batch. (a) Return of episode. (b) Range-normalized return. (c) The range-normalized value of the final round.
    }

    \label{fig:hop}
\end{figure}

A more challenging RL simulation problem is \texttt{Hopper-v4}, based on the Mujoco physics engine. The task is to control a one-legged robot \citep{hop} to hop forward as far as possible. We use the same linear controller code as we used in the previous example \citep{rl_linear}. For \texttt{Hopper-v4}, with 11 state dimensions and 3 action dimensions, this results in a $1 + 3\times 11 = 34$ dimensional controller. See Figure~\ref{fig:hop}.

\subsection{Initialization}
\label{sec:initialization}

The initialization batch is special in that it is the only batch which is designed without the aid of prior measurements. Typically, at this stage, researchers and practitioners will create a space-filling design or randomly sample the parameter space. The rationale is that without measurements, one cannot create a surrogate \citep{hb}. Also, since the location of the maximizer is not know \textit{a priori}, "space-filling designs allow us to build a predictor with better average accuracy" \citep{hb}.

\begin{figure}[bt!]
   \centering
   \includegraphics[width=1.0\linewidth]{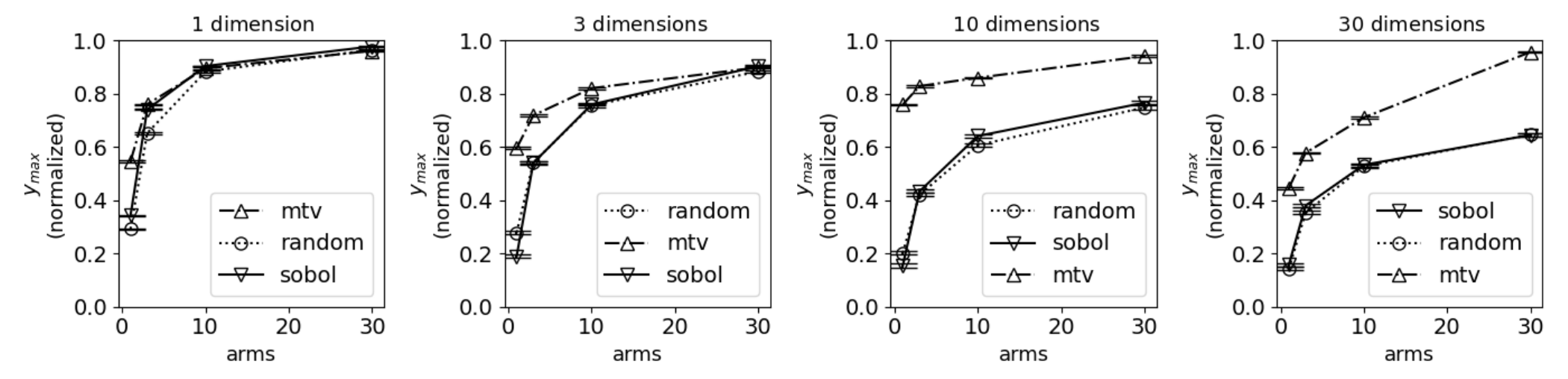}\hfill
    \caption{
    Initializing with MTV finds higher-quality (closer to optimal) parameters than (quasi-)random initialization.
    }
    \label{fig:init}
\end{figure}

Figure~\ref{fig:init} demonstrates, however, that it is possible to reliably locate higher-quality parameters, \textit{de novo}, than one can by uniform sampling. While we can't make predictions about the measurement outcome of a design, once we fix a form for the surrogate (here, a GP) we can make predictions about the epistemic uncertainty (Property 3), i.e. the uncertainty induced by the model. The MTV acquisition function designs a batch that minimizes the post-experiment epistemic uncertainty, and this produces results superior to uniform, random sampling.

We note two features of the plots in Figure~\ref{fig:init}: The impact of switching from random initialization to MTV is greater when the number of arms (the capacity) is lower.

\subsection{Ablations}
\label{sec:ablations}

MTV consists of three crucial components: (i) sampling the points, $x_i \sim p_*(x)$, at which the integral is evaluated, (ii) minimizing the acquisition function, MTV (Equation~\ref{eq:int}), and (iii) selecting initial conditions for the acquisition function optimizer.

\begin{figure}[bt!]
   \centering
   \includegraphics[width=1.0\linewidth]{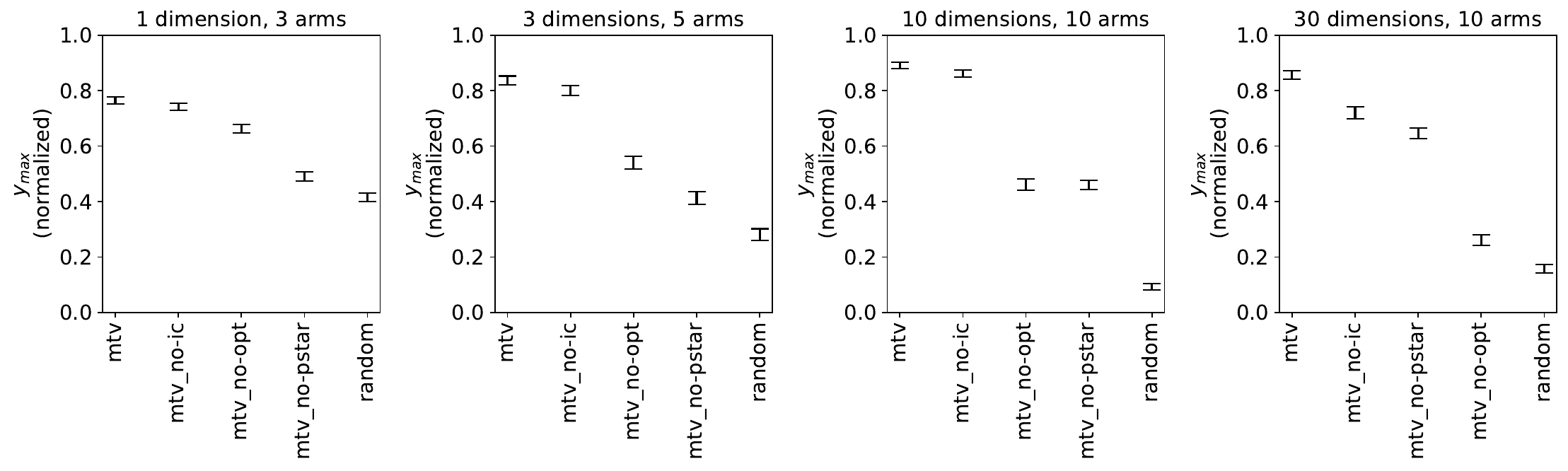}\hfill
    \caption{
    Removing any of the three components of MTV can harm performance. \texttt{mtv\_no\_pstar} replaces $p_*(x)$ samples with Sobol' (uniform) samples.  \texttt{mtv\_no-opt} proposes $p_*(x)$ rather than minimizing MTV, Equation~\ref{eq:int}. \texttt{mtv\_no-ic} starts the acquisition function optimizer at random points instead of samples from $p_*(x)$. 
    }
    \label{fig:ablate}
\end{figure}

Figure~\ref{fig:ablate} compares MTV to ablations that each remove one of the components:
\begin{itemize}
\item \texttt{mtv\_no\_pstar} replaces the samples, $x_i \sim p_*(x)$, in the integral Equation~\eqref{eq:int} with uniformly distributed samples, $x_i \sim \text{Sobol'}$.
\item \texttt{mtv\_no-opt} does not minimize the integral, MTV. Instead it chooses arms as samples from $x_a \sim p_*(x)$ via Algorithm~\ref{alg:mcmc}. Note that an arm sampled from $p_*(x)$ is a Thompson sample \citep{bts}. Approaches similar to \texttt{mtv\_no-opt} were studied in \citet{smcts} and \citet{mcmcts}, in which the authors Thompson-sampled directly from $p_*(x)$ via sequential monte carlo \citep{smcts} and markov-chain monte carlo \citep{mcmcts}.
\item \texttt{mtv\_no-ic} initializes the acquisition function optimizer with uniformly-distributed random samples instead of with samples from $p_*(x)$.
\end{itemize}
Removing any of the above components has a negative impact on performance of \texttt{mtv}. Initializing the acquisition function optimizer carefully (\texttt{mtv\_no-ic}) seems to be more important in higher dimensions, as was also seen in \citet{afic}.

\subsection{Ensembling}
\label{sec:ensembling}
One virtue of MTV is that it presents a unified approach to initialization and improvement, yielding a simpler batch Bayesian optimization procedure. Nevertheless, we know that practitioners will sometimes entertain greater complexity, in search of performance on specific problems, by creating an ensemble of acquisition functions. For example, the function \texttt{gp\_minimize()}, part of \texttt{sciki-optimize}, provides an option \texttt{gp\_hedge}, which creates an ensemble of EI, UCB, and another acquisition function (probability of improvement) \citep{skopt}. Also, in the Black Box Optimization Challenge 2020 \citep{bbc}, all of the top 10 methods used ensembles of acquisition functions and/or surrogate models.

\begin{figure}[bt!]
   \centering
   \includegraphics[width=1.0\linewidth]{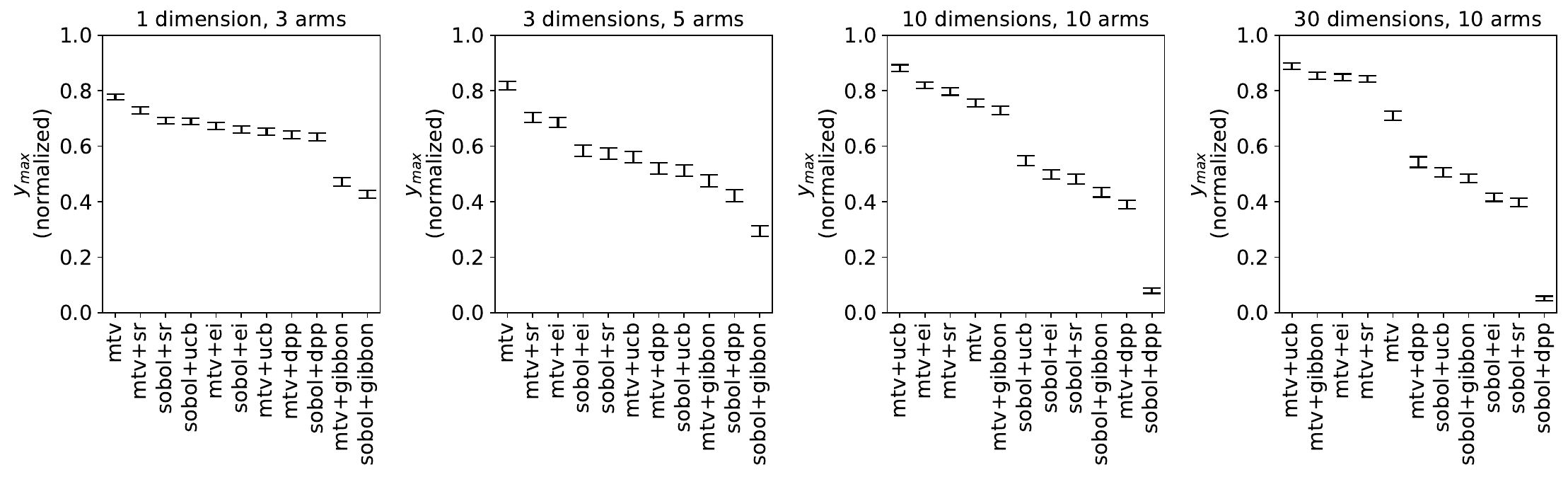}\hfill
    \caption{
    We compare our method, \texttt{mtv}, to ensembles with other acquisition functions, \texttt{mtv+*}, in which the initialization round uses MTV, but the second and third (final) rounds are constructed from the named acquisition function. 
    }
    \label{fig:ensemble}
\end{figure}

In Figure~\ref{fig:ensemble} we compare MTV to a set of ensemble methods that use MTV for the initialization round and other batch acquisition methods for the two subsequent improvement rounds. In some cases the ensembles outperform MTV alone. Although the results are not consistent, as we found in Figure~\ref{fig:compare}, in practice, one may find an ensemble that works well for a specific application.

\section{Conclusion}
We proposed MTV, an acquisition function that is unique in that (i) it generates initialization batch designs via optimization rather than by random sampling, and (ii) it is a single acquisition function that may be used for both initialization and improvement batches, i.e., with or without preexisting measurements. MTV adapts an idea from traditional optimal design of experiments, I-Optimality (IOPT), \citep{iopt,eiopt}, to the batch Bayesian optimization setting. From the DoE perspective, MTV replaces the polynomial surrogate with a GP and weights the IOPT integral, Equation~\ref{eq:int}, to account for pre-existing measurements (if they are available).

MTV's implementation consists of three components: (i) estimation of the acquisition function value (Algorithm~\ref{alg:MTV}) with conditional Gaussian processes, (ii) estimation of the distribution of the maximizer using a novel MCMC algorithm (Algorithm~\ref{alg:mcmc}), and (iii) careful initialization of the acquisition function optimizer to more efficiently locate the desired arms.

Extensive numerical experiments show that MTV outperforms (quasi-)random methods typically used for initialization and outperforms many common batch acquisition functions used to design improvement batches. These results hold across a large range of dimensions. MTV also compares well to other batch Bayesian optimization methods on noisy, reinforcement learning simulators.


\acks{
This work was carried out in affiliation with Yeshiva University. Jiuge Ren
is additionally affiliated with The Goldman Sachs Group, Inc. and David Sweet with DRW Holdings, LLC.
}


\vskip 0.2in
\bibliography{main}

\begin{thebibliography}{52}
\providecommand{\natexlab}[1]{#1}
\providecommand{\url}[1]{\texttt{#1}}
\expandafter\ifx\csname urlstyle\endcsname\relax
  \providecommand{\doi}[1]{doi: #1}\else
  \providecommand{\doi}{doi: \begingroup \urlstyle{rm}\Url}\fi

\bibitem[Adachi et~al.(2023)Adachi, Hayakawa, Hamid, Jørgensen, Oberhauser,
  and Osborne]{sober}
Masaki Adachi, Satoshi Hayakawa, Saad Hamid, Martin Jørgensen, Harald
  Oberhauser, and Michael~A. Osborne.
\newblock Sober: Highly parallel bayesian optimization and bayesian quadrature
  over discrete and mixed spaces.
\newblock \emph{arXiv preprint arXiv:2301.11832}, 2023.

\bibitem[Atkinson(2015)]{iopt}
Anthony~C. Atkinson.
\newblock \emph{Optimal Design}.
\newblock John Wiley \& Sons, Ltd, 2015.
\newblock ISBN 9781118445112.
\newblock \doi{https://doi.org/10.1002/9781118445112.stat04090.pub2}.
\newblock URL
  \url{https://onlinelibrary.wiley.com/doi/abs/10.1002/9781118445112.stat04090.pub2}.

\bibitem[Balandat et~al.(2019)Balandat, Karrer, Jiang, Daulton, Letham, Wilson,
  and Bakshy]{botorch_paper}
Maximilian Balandat, Brian Karrer, Daniel~R. Jiang, Sam Daulton, Benjamin
  Letham, Andrew~Gordon Wilson, and Eytan Bakshy.
\newblock Botorch: Programmable bayesian optimization in pytorch.
\newblock \emph{ArXiv}, abs/1910.06403, 2019.
\newblock URL \url{https://api.semanticscholar.org/CorpusID:204575755}.

\bibitem[Bijl et~al.(2016)Bijl, Schon, van Wingerden, and Verhaegen]{smcts}
Hildo Bijl, Thomas Schon, Jan-Willem van Wingerden, and Michel Verhaegen.
\newblock A sequential monte carlo approach to thompson sampling for bayesian
  optimization.
\newblock \emph{arXiv: Machine Learning}, 2016.
\newblock URL \url{https://api.semanticscholar.org/CorpusID:843648}.

\bibitem[{Binois} et~al.(2021){Binois}, {Collier}, and {Ozik}]{abm}
Mickael {Binois}, Nicholson {Collier}, and Jonathan {Ozik}.
\newblock {A portfolio approach to massively parallel Bayesian optimization}.
\newblock \emph{arXiv e-prints}, art. arXiv:2110.09334, October 2021.
\newblock \doi{10.48550/arXiv.2110.09334}.

\bibitem[Dean et~al.(2017)Dean, Voss, and Draguljic]{exp}
Angela Dean, Daniel Voss, and Danel Draguljic.
\newblock \emph{Design and Analysis of Experiments}.
\newblock Springer, 2017.

\bibitem[Dean et~al.(2020)Dean, Morris, Stufken, and Bingham]{hb}
Angela Dean, Max Morris, John Stufken, and Derek Bingham.
\newblock \emph{Handbo,ok of Design and Analysis of Experiments}.
\newblock Chapman \& Hall, 2020.
\newblock ISBN 9780367570415.
\newblock URL
  \url{https://www.routledge.com/Handbook-of-Design-and-Analysis-of-Experiments/Bingham-Dean-Stufken-Morris/p/book/9780367570415?gclid=CjwKCAjwsKqoBhBPEiwALrrqiHVpBxrR4-dbSm5YtxiE6t6I-DTzgja63PC8YojofdJSP0ObfGAe4xoC-GUQAvD_BwE#}.

\bibitem[dragonfly(2023)]{dragonfly-code}
dragonfly.
\newblock dragonfly, 2023.
\newblock URL
  \url{https://github.com/dragonfly/dragonfly/tree/3eef7d30bcc2e56f2221a624bd8ec7f933f81e40}.

\bibitem[Erez et~al.(2011)Erez, Tassa, and Todorov]{hop}
Tom Erez, Yuval Tassa, and Emanuel Todorov.
\newblock Infinite-horizon model predictive control for periodic tasks with
  contacts.
\newblock In Hugh~F. Durrant-Whyte, Nicholas Roy, and Pieter Abbeel, editors,
  \emph{Robotics: Science and Systems}, 2011.
\newblock ISBN 978-0-262-51779-9.
\newblock URL \url{http://dblp.uni-trier.de/db/conf/rss/rss2011.html#ErezTT11}.

\bibitem[Eriksson and Poloczek(2021)]{scbo}
David Eriksson and Matthias Poloczek.
\newblock Scalable constrained bayesian optimization.
\newblock In Arindam Banerjee and Kenji Fukumizu, editors, \emph{Proceedings of
  The 24th International Conference on Artificial Intelligence and Statistics},
  volume 130 of \emph{Proceedings of Machine Learning Research}, pages
  730--738. PMLR, 13--15 Apr 2021.
\newblock URL \url{https://proceedings.mlr.press/v130/eriksson21a.html}.

\bibitem[Fang and Lin(2003)]{fl}
Kai-Tai Fang and Dennis~K.J. Lin.
\newblock Ch. 4. uniform experimental designs and their applications in
  industry.
\newblock In \emph{Statistics in Industry}, volume~22 of \emph{Handbook of
  Statistics}, pages 131--170. Elsevier, 2003.
\newblock \doi{https://doi.org/10.1016/S0169-7161(03)22006-X}.
\newblock URL
  \url{https://www.sciencedirect.com/science/article/pii/S016971610322006X}.

\bibitem[Foundation(2023)]{gynasium}
Farama Foundation.
\newblock Gymnasium, 2023.
\newblock URL \url{https://github.com/Farama-Foundation/Gymnasium}.

\bibitem[Gardner et~al.(2018)Gardner, Pleiss, Bindel, Weinberger, and
  Wilson]{gpytorch}
Jacob~R. Gardner, Geoff Pleiss, David Bindel, Kilian~Q. Weinberger, and
  Andrew~Gordon Wilson.
\newblock Gpytorch: Blackbox matrix-matrix gaussian process inference with gpu
  acceleration.
\newblock In \emph{Proceedings of the 32nd International Conference on Neural
  Information Processing Systems}, NIPS'18, page 7587–7597, Red Hook, NY,
  USA, 2018. Curran Associates Inc.

\bibitem[Gilks et~al.(1995)Gilks, Richardson, and Spiegelhalter]{mcmc}
W.R. Gilks, S.~Richardson, and D.~Spiegelhalter.
\newblock \emph{Markov Chain Monte Carlo in Practice}.
\newblock Chapman \& Hall/CRC Interdisciplinary Statistics. Taylor \& Francis,
  1995.
\newblock ISBN 9780412055515.
\newblock URL \url{http://books.google.com/books?id=TRXrMWY\_i2IC}.

\bibitem[Gou and Liu(2019)]{mcc_dqn}
Stephen~Zhen Gou and Yuyang Liu.
\newblock Dqn with model-based exploration: efficient learning on environments
  with sparse rewards, 2019.

\bibitem[Gupta et~al.(2019)Gupta, Kohavi, Tang, Xu, Andersen, Bakshy, Cardin,
  Chandran, Chen, Coey, Curtis, Deng, Duan, Forbes, Frasca, Guy, Imbens,
  Saint~Jacques, Kantawala, Katsev, Katzwer, Konutgan, Kunakova, Lee, Lee, Liu,
  McQueen, Najmi, Smith, Trehan, Vermeer, Walker, Wong, and Yashkov]{ab2}
Somit Gupta, Ronny Kohavi, Diane Tang, Ya~Xu, Reid Andersen, Eytan Bakshy,
  Niall Cardin, Sumita Chandran, Nanyu Chen, Dominic Coey, Mike Curtis, Alex
  Deng, Weitao Duan, Peter Forbes, Brian Frasca, Tommy Guy, Guido~W. Imbens,
  Guillaume Saint~Jacques, Pranav Kantawala, Ilya Katsev, Moshe Katzwer, Mikael
  Konutgan, Elena Kunakova, Minyong Lee, MJ~Lee, Joseph Liu, James McQueen,
  Amir Najmi, Brent Smith, Vivek Trehan, Lukas Vermeer, Toby Walker, Jeffrey
  Wong, and Igor Yashkov.
\newblock Top challenges from the first practical online controlled experiments
  summit.
\newblock \emph{SIGKDD Explor. Newsl.}, 21\penalty0 (1):\penalty0 20–35, may
  2019.
\newblock ISSN 1931-0145.
\newblock \doi{10.1145/3331651.3331655}.
\newblock URL \url{https://doi.org/10.1145/3331651.3331655}.

\bibitem[Haftka et~al.(2016)Haftka, Villanueva, and Chaudhuri]{as}
Raphael~T. Haftka, Diane Villanueva, and Anirban Chaudhuri.
\newblock Parallel surrogate-assisted global optimization with expensive
  functions – a survey.
\newblock \emph{JStructural and Multidisciplinary Optimization}, 54:\penalty0
  3--13, 2016.
\newblock URL \url{https://doi.org/10.1007/s00158-016-1432-3}.

\bibitem[Hennig and Schuler(2012)]{es}
Philipp Hennig and Christian~J. Schuler.
\newblock Entropy search for information-efficient global optimization.
\newblock \emph{Journal of Machine Learning Research}, 13\penalty0
  (57):\penalty0 1809--1837, 2012.
\newblock URL \url{http://jmlr.org/papers/v13/hennig12a.html}.

\bibitem[Jiang et~al.(2020)Jiang, Jiang, Balandat, Karrer, Gardner, and
  Garnett]{nonmyopic}
Shali Jiang, Daniel Jiang, Maximilian Balandat, Brian Karrer, Jacob Gardner,
  and Roman Garnett.
\newblock Efficient nonmyopic bayesian optimization via one-shot multi-step
  trees.
\newblock In H.~Larochelle, M.~Ranzato, R.~Hadsell, M.F. Balcan, and H.~Lin,
  editors, \emph{Advances in Neural Information Processing Systems}, volume~33,
  pages 18039--18049. Curran Associates, Inc., 2020.
\newblock URL
  \url{https://proceedings.neurips.cc/paper_files/paper/2020/file/d1d5923fc822531bbfd9d87d4760914b-Paper.pdf}.

\bibitem[Jones et~al.(1998)Jones, Schonlau, and Welch]{ego}
Donald~R Jones, Matthias Schonlau, and William~J. Welch.
\newblock Efficient global optimization of expensive black-box functions.
\newblock \emph{Journal of Global Optimization}, 13:\penalty0 455--492, 1998.
\newblock URL \url{https://doi.org/10.1023/A:1008306431147}.

\bibitem[Kandasamy et~al.(2017)Kandasamy, Krishnamurthy, Schneider, and
  P{\'o}czos]{bts}
Kirthevasan Kandasamy, Akshay Krishnamurthy, Jeff~G. Schneider, and
  Barnab{\'a}s P{\'o}czos.
\newblock Asynchronous parallel bayesian optimisation via thompson sampling.
\newblock \emph{ArXiv}, abs/1705.09236, 2017.
\newblock URL \url{https://api.semanticscholar.org/CorpusID:28218191}.

\bibitem[Kandasamy et~al.(2020)Kandasamy, Vysyaraju, Neiswanger, Paria,
  Collins, Schneider, Poczos, and Xing]{dragonfly-paper}
Kirthevasan Kandasamy, Karun~Raju Vysyaraju, Willie Neiswanger, Biswajit Paria,
  Christopher~R. Collins, Jeff Schneider, Barnabas Poczos, and Eric~P. Xing.
\newblock Tuning hyperparameters without grad students: Scalable and robust
  bayesian optimisation with dragonfly.
\newblock \emph{Journal of Machine Learning Research}, 21\penalty0
  (81):\penalty0 1--27, 2020.
\newblock URL \url{http://jmlr.org/papers/v21/18-223.html}.

\bibitem[Kotthoff et~al.(2021)Kotthoff, Wahab, and Johnson]{ms}
Lars Kotthoff, Hud Wahab, and Patrick Johnson.
\newblock Bayesian optimization in materials science: A survey.
\newblock \emph{ArXiv}, abs/2108.00002, 2021.
\newblock URL \url{https://api.semanticscholar.org/CorpusID:236772166}.

\bibitem[Kudela(2023)]{centerbias}
Jakub Kudela.
\newblock The evolutionary computation methods no one should use, 2023.

\bibitem[Letham and Bakshy(2019)]{fb3}
Benjamin Letham and Eytan Bakshy.
\newblock Bayesian optimization for policy search via online-offline
  experimentation.
\newblock \emph{Journal of Machine Learning Research}, 20\penalty0
  (145):\penalty0 1--30, 2019.
\newblock URL \url{http://jmlr.org/papers/v20/18-225.html}.

\bibitem[Letham et~al.(2017)Letham, Karrer, Ottoni, and Bakshy]{qnei}
Benjamin Letham, Brian Karrer, Guilherme Ottoni, and Eytan Bakshy.
\newblock Constrained bayesian optimization with noisy experiments.
\newblock \emph{ArXiv}, abs/1706.07094, 2017.
\newblock URL \url{https://api.semanticscholar.org/CorpusID:37601814}.

\bibitem[Letham et~al.(2019)Letham, Karrer, Ottoni, and Bakshy]{fb}
Benjamin Letham, Brian Karrer, Guilherme Ottoni, and Eytan Bakshy.
\newblock Constrained bayesian optimization with noisy experiments.
\newblock \emph{Bayesian Analysis}, 14\penalty0 (2):\penalty0 495--519, 2019.
\newblock \doi{10.1214/18-BA1110}.
\newblock URL \url{https://doi.org/10.1214/18-BA1110}.

\bibitem[Li and Deng(2021)]{eiopt}
Yiou Li and Xinwei Deng.
\newblock An efficient algorithm for elastic i-optimal design of generalized
  linear models.
\newblock \emph{Canadian Journal of Statistics}, 49\penalty0 (2):\penalty0
  438--470, 2021.
\newblock \doi{https://doi.org/10.1002/cjs.11571}.
\newblock URL \url{https://onlinelibrary.wiley.com/doi/abs/10.1002/cjs.11571}.

\bibitem[Mania et~al.(2018)Mania, Guy, and Recht]{rl_linear}
Horia Mania, Aurelia Guy, and Benjamin Recht.
\newblock Simple random search provides a competitive approach to reinforcement
  learning, 2018.

\bibitem[Meta(2023{\natexlab{a}})]{ax}
Meta.
\newblock Ax, 2023{\natexlab{a}}.
\newblock URL \url{https://ax.dev}.

\bibitem[Meta(2023{\natexlab{b}})]{botorch_code}
Meta.
\newblock Botorch, 2023{\natexlab{b}}.
\newblock URL \url{https://botorch.org}.

\bibitem[Moore(1990)]{mcc_orig}
Andrew Moore.
\newblock Efficient memory-based learning for robot control.
\newblock Technical report, Carnegie Mellon University, Pittsburgh, PA,
  November 1990.

\bibitem[Moss et~al.(2021)Moss, Leslie, Gonzalez, and Rayson]{gibbon}
Henry~B. Moss, David~S. Leslie, Javier Gonzalez, and Paul Rayson.
\newblock Gibbon: General-purpose information-based bayesian optimisation.
\newblock \emph{Journal of Machine Learning Research}, 22\penalty0
  (235):\penalty0 1--49, 2021.
\newblock URL \url{http://jmlr.org/papers/v22/21-0120.html}.

\bibitem[Nava et~al.(2022)Nava, Mutny, and Krause]{dpp}
Elvis Nava, Mojmir Mutny, and Andreas Krause.
\newblock Diversified sampling for batched bayesian optimization with
  determinantal point processes.
\newblock In Gustau Camps-Valls, Francisco J.~R. Ruiz, and Isabel Valera,
  editors, \emph{Proceedings of The 25th International Conference on Artificial
  Intelligence and Statistics}, volume 151 of \emph{Proceedings of Machine
  Learning Research}, pages 7031--7054. PMLR, 28--30 Mar 2022.
\newblock URL \url{https://proceedings.mlr.press/v151/nava22a.html}.

\bibitem[NIST/SEMATECH(2012)]{nist}
NIST/SEMATECH.
\newblock Nist/sematech e-handbook of statistical methods, 2012.
\newblock URL
  \url{https://www.itl.nist.gov/div898/handbook/pri/section5/pri53.htm}.
\newblock Accessed: 2023-12-10.

\bibitem[Quin et~al.(2023)Quin, Weyns, Galster, and Silva]{ab}
Federico Quin, Danny Weyns, Matthias Galster, and Camila~Costa Silva.
\newblock A/b testing: A systematic literature review.
\newblock \emph{ArXiv}, abs/2308.04929, 2023.
\newblock URL \url{https://api.semanticscholar.org/CorpusID:260735919}.

\bibitem[Ren et~al.(2018)Ren, Ward, Williams, Laws, Wolverton,
  Hattrick-Simpers, and Mehta]{ms2}
Fang Ren, Logan Ward, Travis Williams, Kevin~J. Laws, Christopher Wolverton,
  Jason Hattrick-Simpers, and Apurva Mehta.
\newblock Accelerated discovery of metallic glasses through iteration of
  machine learning and high-throughput experiments.
\newblock \emph{Science Advances}, 4\penalty0 (4):\penalty0 eaaq1566, 2018.
\newblock \doi{10.1126/sciadv.aaq1566}.
\newblock URL \url{https://www.science.org/doi/abs/10.1126/sciadv.aaq1566}.

\bibitem[Santner et~al.(2019)Santner, Williams, and Notz]{dace}
Thomas~J. Santner, Brian~J. Williams, and William~I. Notz.
\newblock \emph{The Design and Analysis of Computer Experiments}.
\newblock Springer New York, NY, 2019.
\newblock ISBN 9781493988471.
\newblock \doi{https://doi.org/10.1007/978-1-4939-8847-1}.
\newblock URL \url{https://link.springer.com/book/10.1007/978-1-4939-8847-1}.

\bibitem[Scikit-Optimize(2023)]{skopt}
Scikit-Optimize.
\newblock scikit-optimize, 2023.
\newblock URL \url{https://scikit-optimize.github.io}.

\bibitem[SciPy(2023)]{scipy}
SciPy.
\newblock scipy, 2023.
\newblock URL \url{https://scipy.org}.

\bibitem[Speagle(2019)]{mcmcintro}
Joshua~S. Speagle.
\newblock A conceptual introduction to markov chain monte carlo methods.
\newblock \emph{arXiv: Other Statistics}, 2019.
\newblock URL \url{https://api.semanticscholar.org/CorpusID:203591670}.

\bibitem[Srinivas et~al.(2010)Srinivas, Krause, Kakade, and Seeger]{ucb}
Niranjan Srinivas, Andreas Krause, Sham Kakade, and Matthias Seeger.
\newblock Gaussian process optimization in the bandit setting: No regret and
  experimental design.
\newblock In \emph{Proceedings of the 27th International Conference on
  International Conference on Machine Learning}, ICML'10, page 1015–1022,
  Madison, WI, USA, 2010. Omnipress.
\newblock ISBN 9781605589077.

\bibitem[Surjanovic and Bingham(2013)]{optfun}
Sonya Surjanovic and Derek Bingham.
\newblock Virtual library of simulation experiments: Optimization test
  problems.
\newblock \url{http://www.sfu.ca/~ssurjano/optimization.html}, 2013.
\newblock Accessed: 2023-10-25.

\bibitem[Sweet(2023)]{e4e}
David Sweet.
\newblock \emph{Experimentation for Engineers}.
\newblock Manning, 2023.
\newblock ISBN 9781617298158.
\newblock URL
  \url{https://www.manning.com/books/experimentation-for-engineers}.

\bibitem[Turner et~al.(2021)Turner, Eriksson, McCourt, Kiili, Laaksonen, Xu,
  and Guyon]{bbc}
Ryan Turner, David Eriksson, Michael McCourt, Juha Kiili, Eero Laaksonen, Zhen
  Xu, and Isabelle Guyon.
\newblock Bayesian optimization is superior to random search for machine
  learning hyperparameter tuning: Analysis of the black-box optimization
  challenge 2020.
\newblock In Hugo~Jair Escalante and Katja Hofmann, editors, \emph{Proceedings
  of the NeurIPS 2020 Competition and Demonstration Track}, volume 133 of
  \emph{Proceedings of Machine Learning Research}, pages 3--26. PMLR, 06--12
  Dec 2021.
\newblock URL \url{https://proceedings.mlr.press/v133/turner21a.html}.

\bibitem[Wang(2020)]{tut}
Jie Wang.
\newblock An intuitive tutorial to gaussian processes regression.
\newblock \emph{ArXiv}, abs/2009.10862, 2020.
\newblock URL \url{https://api.semanticscholar.org/CorpusID:221857371}.

\bibitem[Wang et~al.(2023)Wang, Jin, Schmitt, and Olhofer]{rec}
Xilu Wang, Yaochu Jin, Sebastian Schmitt, and Markus Olhofer.
\newblock Recent advances in bayesian optimization.
\newblock \emph{ACM Comput. Surv.}, 55\penalty0 (13s), jul 2023.
\newblock ISSN 0360-0300.
\newblock \doi{10.1145/3582078}.
\newblock URL \url{https://doi.org/10.1145/3582078}.

\bibitem[Wang and Jegelka(2017)]{mves}
Zi~Wang and Stefanie Jegelka.
\newblock Max-value entropy search for efficient bayesian optimization.
\newblock In \emph{Max-Value Entropy Search for Efficient Bayesian
  Optimization}, ICML'17, page 3627–3635. JMLR.org, 2017.

\bibitem[Wilson et~al.(2017{\natexlab{a}})Wilson, Moriconi, Hutter, and
  Deisenroth]{reparam}
James~T. Wilson, Riccardo Moriconi, Frank Hutter, and Marc~P. Deisenroth.
\newblock The reparameterization trick for acquisition functions.
\newblock In \emph{NIPS Workshop on Bayesian Optimization}, 2017{\natexlab{a}}.
\newblock URL \url{https://bayesopt.github.io/papers/2017/32.pdf}.

\bibitem[Wilson et~al.(2017{\natexlab{b}})Wilson, Moriconi, Hutter, and
  Deisenroth]{q}
James~T. Wilson, Riccardo Moriconi, Frank Hutter, and Marc~Peter Deisenroth.
\newblock The reparameterization trick for acquisition functions.
\newblock \emph{ArXiv}, abs/1712.00424, 2017{\natexlab{b}}.
\newblock URL \url{https://api.semanticscholar.org/CorpusID:22554301}.

\bibitem[Yi et~al.(2024)Yi, Wei, Cheng, He, and Sui]{mcmcts}
Zeji Yi, Yunyue Wei, Chu~Xin Cheng, Kaibo He, and Yanan Sui.
\newblock Improving sample efficiency of high dimensional bayesian optimization
  with mcmc, 2024.

\bibitem[Zhao et~al.(2023)Zhao, Yang, Qiu, and Wang]{afic}
Jiayu Zhao, Renyu Yang, Shenghao Qiu, and Zheng Wang.
\newblock Enhancing high-dimensional bayesian optimization by optimizing the
  acquisition function maximizer initialization.
\newblock \emph{ArXiv}, abs/2302.08298, 2023.
\newblock URL \url{https://api.semanticscholar.org/CorpusID:256900880}.

\end{thebibliography}

\end{document}